\documentclass[journal]{IEEEtran}
\ifCLASSINFOpdf
\else
\fi

\usepackage{cite}
\usepackage{graphicx}
\usepackage{algorithmic}
\usepackage{algorithm}
\usepackage{amsmath}
\usepackage{MnSymbol}%
\usepackage{booktabs}
\usepackage{color}
\usepackage{threeparttable}
\usepackage{array}

\AtBeginDocument{%
	\providecommand\BibTeX{{%
			\normalfont B\kern-0.5em{\scshape i\kern-0.25em b}\kern-0.8em\TeX}}}

\usepackage{multirow}

\usepackage{enumitem}

\usepackage{tikz}

\def\checkmark{\tikz\fill[scale=0.4](0,.35) -- (.25,0) -- (1,.7) -- (.25,.15) -- cycle;} 

\graphicspath{{figure/}}

\begin{document}
\bstctlcite{IEEEexample:BSTcontrol}
\title{Contrastive Learning of Person-independent Representations for Facial Action Unit Detection}
%
%
%

\author{Yong Li,
	Shiguang Shan,~\IEEEmembership{Fellow,~IEEE}
	\IEEEcompsocitemizethanks{\IEEEcompsocthanksitem 
		Corresponding author: Shiguang Shan, sgshan@ict.ac.cn
		\protect\\
		Yong Li is with the Key Laboratory of Intelligent Perception and Systems for High-Dimensional Information, Ministry of Education, School of Computer Science and Engineering, Nanjing University of Science and Technology, Nanjing, 210094, China (e-mail: yong.li@njust.edu.cn)
		\IEEEcompsocthanksitem Shiguang Shan is with the Key Laboratory of Intelligent Information Processing of Chinese Academy of Sciences, Institute of Computing Technology, CAS, Beijing 100190, China, and with the University of Chinese Academy of Sciences, Beijing 100049, China, and also with Peng Cheng Laboratory, Shenzhen, 518055, China (e-mail: sgshan@ict.ac.cn).}
		}

\markboth{Journal of \LaTeX\ Class Files,~Vol.~14, No.~8, August~2015}%
{Shell \MakeLowercase{\textit{et al.}}: Bare Demo of IEEEtran.cls for IEEE Journals}

\maketitle

\begin{abstract}
Facial action unit (AU) detection, aiming to classify AU present in the facial image, has long suffered from insufficient AU annotations. In this paper, we aim to mitigate this data scarcity issue by learning AU representations from a large number of unlabelled facial videos in a contrastive learning paradigm. We formulate the self-supervised AU representation learning signals in two-fold: (1) AU representation should be frame-wisely discriminative within a short video clip; (2) Facial frames sampled from different identities but show analogous facial AUs should have consistent AU representations. As to achieve these goals, we propose to contrastively learn the AU representation within a video clip and devise a cross-identity reconstruction mechanism to learn the person-independent representations. Specially, we adopt a margin-based temporal contrastive learning paradigm to perceive the temporal AU coherence and evolution characteristics within a clip that consists of consecutive input facial frames.  Moreover, the cross-identity reconstruction mechanism facilitates pushing the faces from different identities but show analogous AUs close in the latent embedding space. Experimental results on three public AU datasets demonstrate that the learned AU representation is discriminative for AU detection. Our method outperforms other contrastive learning methods and significantly closes the performance gap between the self-supervised and supervised AU detection approaches.

\end{abstract}

\IEEEpeerreviewmaketitle

\section{Introduction}
\label{sec:introduction}

\label{sec:intro}

\begin{figure}[htb]
	\includegraphics[width=1.0\linewidth]{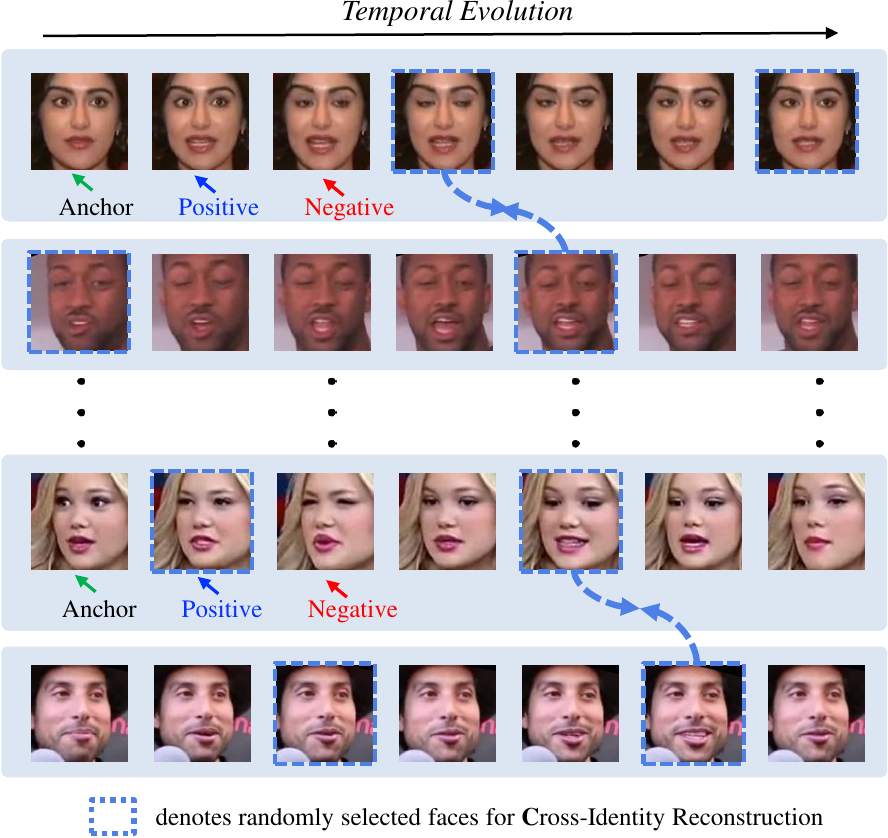}
	\caption{
		Main idea of our proposed contrastively learning the person-independent representations for AU detection method (CLP). CLP learns the frame-wisely discriminative AU representations via temporally contrastive learning. To further remove the person-specific nuisances, CLP exploits cross-identity reconstruction mechanism to push the faces from different identities but show consistent AUs close in the latent embedding space, thus encoding the inter-identity consistency in the representations.
	}
	\label{fig:main_idea}
\end{figure}

\IEEEPARstart{F}{acial} Action Coding System (FACS) \cite{ekman1978manual} is a systematic approach to describe what a face looks like when facial muscle movements have occurred. It describes the presence and intensity of various facial movements. In FACS, each facial action unit (AU) is hypothesized to correspond to the contraction of a distinct facial muscle or a distinct grouping of muscles that are visible as a specific facial movement \cite{martinez2017automatic}. For example, the raising of the inner corners of the eyebrows corresponds to AU1. Lowering the inner corners of the brows means the activation of AU4. With FACS, each facial expression can be expressed via a combination of multiple AUs. Therefore, a reliable AU detection system is of great importance for the analysis of facial expressions. Furthermore, AU detection technology also holds promise to an abundance of practical applications, such as affect analysis \cite{kotsia2008texture}, automatic pain estimation \cite{zafar2014pain}, and human-computer interaction \cite{bartlett2003real}.

In recent years, the development of AU detection has gained considerable improvements based on the progress in deep learning techniques\cite{zhao2016deep, li2017eac, corneanu2018deep, shao2018deep, li2019semantic, yang2021exploiting, song2021hybrid}. To learn the AU-specific representation and improve the AU detection performance, many of the prior works typically adopt the facial landmarks to localize or crop the AU-related facial regions. Despite the improvements, people have realized that such deep learning-based supervised methods are data starved. 
In fact, existing AU datasets like BP4D \cite{zhang2013high} and DISFA \cite{mavadati2013disfa} only comprise limited subjects as well as the facial images. As a consequence, the  AU detection models trained on these datasets usually overfit on person-specific features and can be poorly generalized \cite{girard2015much, zhao2018learning}. 
However, the acquisition of AU annotations is intellectually expensive, error-prone, and cumbersome  \cite{zhao2018learning}. Human coders need to be trained for many weeks to reliably identify if a specific AU is activated or not. It has been verified that an AU expert needs to take almost half an hour to manually annotate one AU for a one-minute video \cite{zhao2018learning,sun2021emotion}.

To alleviate the demand for a large amount and accurate AU annotations, a number of literatures propose to exploit the semi-/weakly-supervised methods to learn more generalizable AU representations. Among them, the former \cite{zhao2018learning} incorporates both the labelled and unlabelled AU data by assuming the faces to be clustered by AUs. The latter seeks to leverage the facial images with incomplete or noisy AU annotations \cite{zhang2018weakly,peng2018weakly}. Different with them, we challenge the AU data scarcity issue via learning the discriminative AU representations from the practically infinite amount of unlabelled facial videos in a self-supervised contrastive learning paradigm. Intuitively, the frame sequences within a video naturally offer the temporal evolution and consistency characteristics for the video objects, which have been extensively utilized as self-supervisory signals \cite{kong2020cycle, Miech_2020_CVPR, li2021intra, wu2021contrastive, Qian_2021_CVPR}.
However, the recent pervasive video-based contrastive learning methods usually learn video-level representation \cite{kong2020cycle, li2021intra, Miech_2020_CVPR, li2021intra} or frame correspondence \cite{wang2019learning, wu2021contrastive}, while AU representation should be frame-wisely discriminative within a video clip \cite{yang2019facs3d, tellamekala2019temporally} and consistent across identities that show analogous AUs \cite{almaev2015learning, song2021self}.

To effectively learn the discriminative AU representations, we propose to \textbf{C}ontrastively \textbf{L}earn the \textbf{P}erson-independent (CLP) representations for facial AU detection. Fig.~\ref{fig:main_idea} illustrates the main idea of the proposed CLP.
We sample a consecutive frame sequence from a video clip where the AUs show temporally coherence and evolution naturally.
To perceive the frame-wisely distinctiveness of the consecutive faces, CLP is tasked to learn their feature similarity and distinctiveness via temporally contrastive learning. 
As shown in the first frame sequence in Fig.~\ref{fig:main_idea}, given an anchor frame in a video sequence, the subsequent frames naturally contain abundant positive and negative samples that can be used to construct triplets according to their temporal distance to the anchor frame. To make use of this nature,  we introduce a temporally contrastive learning method that carefully takes the temporal coherence and evolution dynamics into consideration, which will be carefully described in  Sec.~\ref{sec:temporal_contrastive_learning}. 

In order to facilitate learning consistent AU representations across identities that illustrate analogous AUs, we generalize the contrastive learning with positive image pairs across different facial videos. For an arbitrary facial frame sampled from a subject, CLP reconstructs its AU representation via re-weighing the representations of the training samples from other identities.
We hypothesize this inter-subject invariance learning paradigm can potentially perceive the high-level AU semantics beyond the undesired subject-dependent nuisances. As shown in Fig.~\ref{fig:main_idea}, the cross-identity facial frames with consistent AUs will be pulled close in the latent embedding space. Concretely, given a facial frame $q$ randomly sampled from a video, we firstly generate a copy of $q$ as $k$ via artificial pixel-level augmentation, then approximate $q$ by composing frames sampled from other identities that illustrate similar AUs with $q$. Finally, we compare the composed frame $\hat{q}$ with $k$ for cross-identity reconstruction (CIR). However, CIR may not be achievable due to limited video sequences within a training batch. To obtain a dictionary that consists of a large amount of AU representations from different identities, we build a large and consistent dictionary as a queue, with the sequences in the current mini-batch enqueued and the oldest mini-batch dequeued, decoupling it from the mini-batch size \cite{he2020momentum}. We present the details of how we learn person-independent facial AU representations in  Sec.~\ref{sec:Cross_video_Cycle_Consistency}. Compared with the vanilla weakly-/semi-supervised methods, the proposed CLP can learn discriminative AU features without AU labels and is regardless of the assumptions on label distribution.

In summary, the contributions of this paper can be concluded as follows:
\begin{itemize}
	\item We formulate a contrastive self-supervised learning method for encoding the discriminative AU representations from the practically infinite amount of unlabelled facial videos. Given a common observation that a facial video usually comes from only a single subject, the desired AU representations should be discriminative within a video clip and consistent across different identities that show analogous facial AUs.
	\item We propose to contrastively encode the intra-video distinctiveness of the AU representations via deriving the self-supervised pseudo signals  from the temporal characteristics within a short video clip. Furthermore, the representations are optimized to be person-independent with CIR. To the best of our knowledge, our proposed CLP is the first work that uses the self-supervised CIR for AU representation learning.
	\item Experimental results show CLP outperforms other self-supervised contrastive learning methods and significantly closes the performance gap between the self-supervised and supervised AU detection approaches.
\end{itemize}

\section{Related Work}

\subsection{Facial action unit detection}
AU detection has been studied for decades and many methods have been proposed \cite{martinez2017automatic}. To achieve the promising AU detection performance, different hand-crafted features have been introduced to encode the characteristics of AUs, such as Histogram of Oriented
Gradient (HOG) \cite{yang2019facial}, local binary pattern (LBP) \cite{sandbach2012local}, Gabor \cite{fabian2016emotionet} etc. Recently, AU detection has achieved considerable improvements due to development of deep learning and convolutional neural network (CNN). 
Since AU corresponds to the movements of regional facial muscles, many methods aim to detect the occurrence of AU based on location \cite{zhao2016deep,li2017eac,li2017action,shao2018deep}. 
For example, Zhao \textit{et al.} \cite{zhao2016deep} trained multiple region-specific convolutional filters from the face's sub-areas using a regionally connected convolutional layer.
EAC-Net \cite{li2017eac} and ROI \cite{li2017action} developed to extract facial AU features around the manually defined facial landmarks that are  resilient to non-rigid facial shape changes.  SEV-Net \cite{yang2021exploiting} utilized the AU semantic description as auxiliary information for AU detection. AU-RCNN \cite{ma2019r} encoded the prior expert knowledge into the local facial regions for precise AU detection. To avoid predefining the fixed attentions manually, ARL \cite{shao2019facial} proposed to learn  channel-wise and spatial attentions for AU detection in an end-to-end manner,  Recently, Jacob \textit{et al.} \cite{jacob2021facial} used a transformer-based encoder to capture the relationships between AUs. However, these supervised methods rely on precisely annotated images and often overfit on a specific dataset as a result of insufficient training images. Specially, AU detection should be independent of any specific
subject, it is necessary as well as imperative to remove the identity-specific effects on the AU detection model.

Several literature proposed to learn the subject-independent AU features. Among them, Zhang \textit{et al.} \cite{zhang2018identity} proposed the Adversarial Training Framework (ATF). ATF uses the adversarial training paradigm to make the learned features effective for AU detection and invariant to subject variations by minimizing the AU loss while maximizing the face recognition loss. Almaev \textit{et al.} \cite{almaev2015learning} constructed person-specific models for AU detection via transferring the knowledge of the easily elicited AUs to those of hardly elicited ones. Song \textit{et al.} \cite{song2021self} proposed to learn person-specific facial dynamics with a two-step training strategy. In the second phase, a set of intermediate filters were exclusively optimized to represent the person-specific facial dynamics with the person-specific videos. Tu \textit{et al.} \cite{tu2019idennet} used the AU- and identity-annotated datasets that contain numerous subjects to extract identity-dependent image features, and performed the identity-aware AU Detection in the test phase. SO-Net \cite{yang2020set} augmented the training data by adding set operations to both the feature and label spaces. The features are thus generalizable to unseen subjects.
Different with these supervised methods that aim to eliminate the person-specific effects, our proposed CLP aims to learn person-independent AU representations from a large amount of unlabeled facial videos via cross-identity reconstruction.

Recently, weakly-supervised \cite{zhao2018learning,peng2018weakly,zhang2018weakly} and self-supervised \cite{wiles2018self,lu2020self,li2019self,li2020learning} methods have attracted a lot of attention to mitigate the AU data scarcity issue.
Weakly supervised methods exploit noisy or incomplete AU annotations\cite{zhao2018learning}. They usually learn AU classifiers from the domain knowledge\cite{zhang2018weakly,peng2018weakly}. Besides,  MAL \cite{li2021meta} proposed to enhance the AU detection task with auxiliary labelled facial expression (FE) images. Specially, MAL automatically selects the highly related samples by learning adaptive weights for the training FE samples in a meta learning manner. 
For the self-supervised methods, Lu \textit{et al.} \cite{lu2020self} proposed an aggregate ranking loss that learns AU representations by encoding the temporal consistency.
Fab-Net \cite{wiles2018self} was optimized to map a source frame to a target frame by predicting a flow field between them. TCAE \cite{li2019self} and TAE \cite{li2020learning} proposed to learn the pose-invariant AU representation via predicting separate displacements for pose and AU. EmoCo \cite{sun2021emotion} learns the AU representation via classifying the features into different facial expression categories and contrasting features within each emotional category in the instance level. Different from them, we propose to learn the intra-video discriminative and inter-video invariant AU representations contrastively from a large amount of unlabeled videos.

\subsection{Self-supervised contrastive learning methods}
Recently, self-supervised representation learning approaches based on contrastive learning paradigm in the latent space have
shown great promise, achieving state-of-the-art results \cite{wu2018unsupervised,chen2020simple,he2020momentum,chen2021exploring,grill2020bootstrap,jing2020self,zbontar2021barlow}. 
Typically, self-supervised learning methods create the self-defined pseudo labels as supervision and learn representations, which are then used in downstream tasks.
These methods usually follow an instance discrimination task: a query matches a key if they are the encoded views of the same image under different transformations. Using this pretext task, Chen \textit{et al.} \cite{chen2020simple} introduced a simple framework (SimCLR), where a base encoder network and a projection head are trained to maximize the agreements using a contrastive loss.
He \textit{et al.} \cite{he2020momentum} proposed MoCo that contains a momentum network to store a queue of a large number of negative samples for efficient contrastive learning. Chen \textit{et al.} \cite{chen2021exploring} learned general representations without negative sample pairs and avoids the undesired collapsing solution via a stop-gradient operation. These contrastive learning methods learn the general representation which might not be AU discriminative because they do not encode the temporal consistency and are not cross-video consistent, we will compare them with CLP in Sec.~\ref{sec:Comparison_with_the state-of-the-art}. 

Going beyond learning from a single image, an unlabeled video sequence naturally contains temporal coherence and evolution characteristics \cite{yao2020seco}, thus temporal information is a natural supervision signal for self-supervised learning from videos. Recent works generalize the similarity learning between video frames under the contrastive learning paradigm and learn meaningful video-level representations \cite{kong2020cycle,Miech_2020_CVPR,li2021intra,Qian_2021_CVPR,sun2021cross,wu2021contrastive}. However, the learned representations of these general self-supervised contrastive learning approaches are not a dedicated AU representation. It is also controversial that a universal representation exists for all the downstream tasks.

\begin{figure*}[h]
	\centering
	\includegraphics[width=0.9\linewidth]{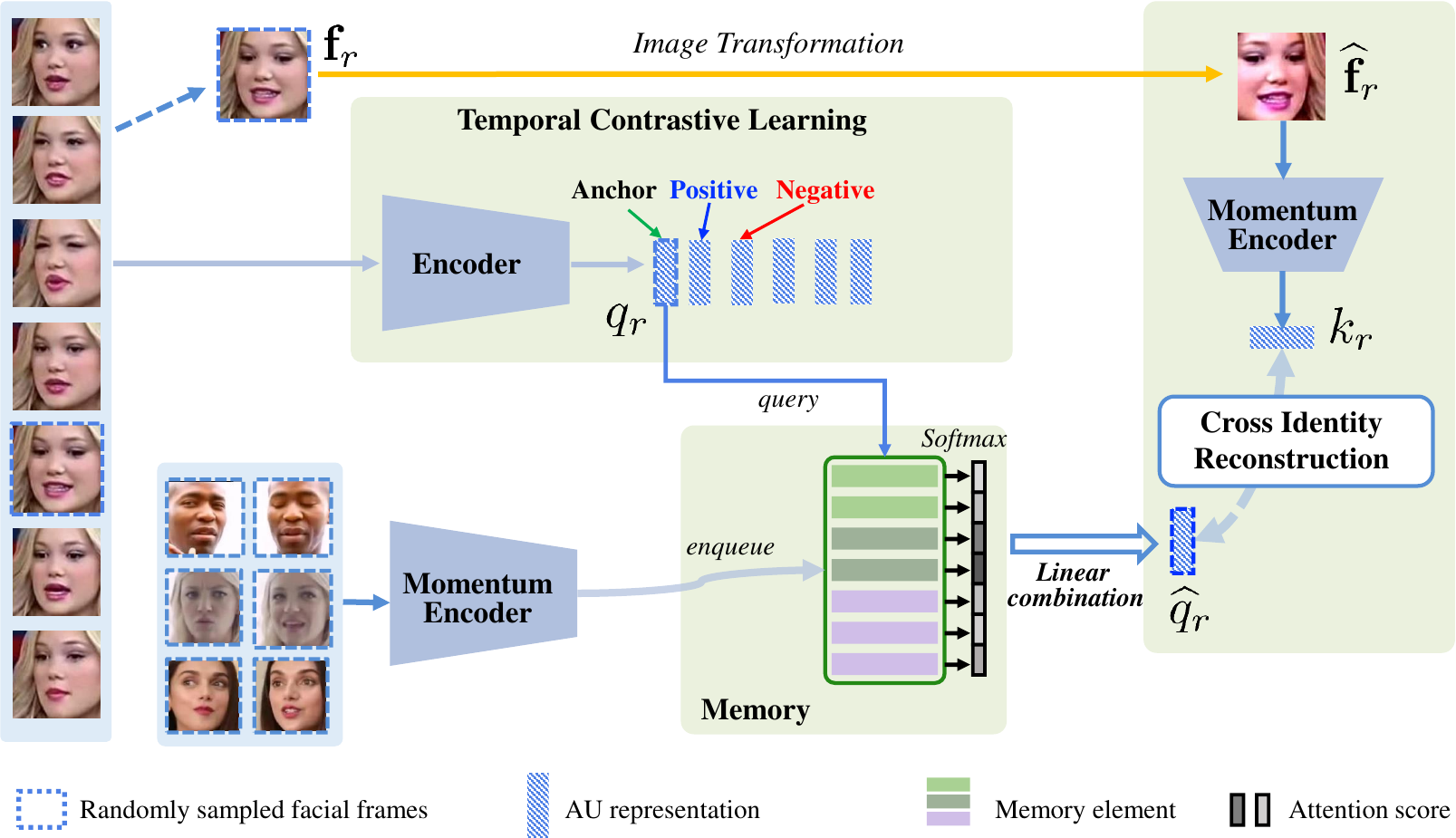}
	\caption{
		The framework of CLP. To make the representations distinctive within a short video clip, we randomly sample a temporally consecutive sequence and construct a set of triplets according to the temporal interval for intra-video contrastive learning (Sec.~\ref{sec:temporal_contrastive_learning}). Besides, we exploit the cross-identity reconstruction (CIR) mechanism to make the representations consistent for faces of different subjects that show analogous AUs. In CIR, we reconstruct the soft nearest neighbour $\hat{q}_{r}$ of $q_{r}$ from a dictionary, which will be described in Sec.~\ref{sec:Cross_video_Cycle_Consistency}.
	}
	\label{fig:framework_overview}
\end{figure*}

\section{Temporal Contrastive learning and Cross-identity Reconstruction in CLP}

In this paper, we exploit the key insight that facial AU representations should encode the temporal coherence and dynamics within a video clip and should contain less identity information. Thus, we devise a natural source of ``self'' supervision to exploit a large amount of unlabeled facial videos.
An overview of the training procedure of the proposed CLP is presented in Fig.~\ref{fig:framework_overview}.
The goal of CLP is to learn the frame-wise AU representations that are distinctive within a video clip and consistent between faces of different subjects that show analogous AUs.
To achieve the first goal, i.e., encoding the distinctiveness of the AU representations within a video clip,  we train the proposed CLP with the temporally consecutive sequences for \textit{intra-video} contrastive learning. To further eliminate the person-specific effects in the AU representations, we randomly sample discontinuous facial frames from different subjects and perform cross identity reconstruction, thus extending the intra-video contrastive learning to \textit{inter-video} invariance learning.

\begin{figure*}[h]
	\centering
	\includegraphics[width=0.8\linewidth]{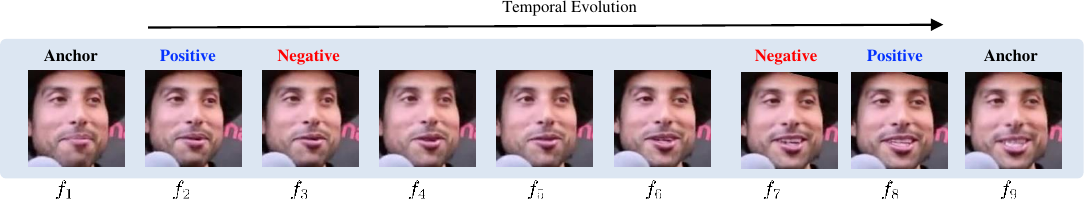}
	\caption{
		An example facial sequence with nine frames sampled from the Voxceleb2 dataset \cite{chung2018voxceleb2}.
		We manually construct a set of triplets: $(f_{1}^a, f_{2}^p, f_{3}^n), (f_{1}^a, f_{3}^p, f_{4}^n), \cdots, (f_{1}^a, f_{8}^p, f_{9}^n)$. This intra-video temporal contrastive learning paradigm naturally learns to classify the frames based on their temporal distance from an anchor frame. Another set of triplets can be constructed in the reversed temporal order.
	}
	\label{fig:frame_sequence}
\end{figure*}

\subsection{Temporal Contrastive learning in CLP}
\label{sec:temporal_contrastive_learning}

Given a common observation that a facial video usually comes from only a single subject \cite{chung2018voxceleb2, song2021self}, we are capable of exploiting the temporal coherence and evolution characteristics to derive the self-supervised contrastive learning signals for AU representations learning.  Inside a consecutive facial sequence, the AU representations of the continuous facial frames should be closer than that of the non-continuous ones. Inspire by this intuition, we use the temporal distance to  measure the proximities of the AU representations of the unlabeled facial sequences. Since our goal is to train a network that learns frame-wise AU representation, our proposed CLP is tasked to rank the intra-video facial frames based on their temporal positions. This way CLP is forced to perceive the frame-wise distinctiveness of the consecutive facial frames within any facial video clips.

Formally, let $\mathcal{V} = \{\mathbf{V}_i, 1 \leq i \leq \text{I}\}$ be the unlabeled training video set, where $i$ denotes the index of the facial videos, $\text{I}$ is the total number of videos. A randomly sampled consecutive facial sequence $\mathbf{F}_i$ from $i$-th video can thus be denoted as:  $\mathbf{F}_i = \{\mathbf{f}_{j},  1 \leq j \leq \text{J}\}$, where  $j$ means the frame index, $\text{J}$ denotes the total number of frames in $\mathbf{F}_i$. As exemplified in Fig.~\ref{fig:frame_sequence}, we adopt the temporal order as the self-supervised pseudo signal to rank the facial frames according to the proximities of their AU representations. As temporal AU changes are quite minor, the changes of facial displays between two sibling frames might be too subtle to be perceived. Thus, the facial frames in a consecutive sequence can be sampled with a stride of $\mathbf{s}$ to enlarge the temporal differences. $\mathbf{s}$ will be discussed in Sec.~\ref{sec:experimental_details}. With a proper sampling stride, the sequential faces in  Fig.~\ref{fig:frame_sequence} illustrate progressive temporal changes of the AUs. Afterwards, CLP constructs a set of facial triplets $ \{\mathbf{T}_t\}_{t=1}^{|\mathbf{T}|}$ for each facial sequence, where each triplet $\mathbf{T}_t = \{\mathbf{f}^a, \mathbf{f}^p, \mathbf{f}^n \}$ consists of an anchor frame $\mathbf{f}^a$, a positive frame $\mathbf{f}^p$ and a negative frame $\mathbf{f}^n$. Subsequently, CLP makes $\mathbf{f}^p$ closer to  $\mathbf{f}^a$ than $\mathbf{f}^n$ by at least a margin $m$ in the normalized embedding space. With a large number of unlabeled  videos,  we are capable of constructing abundant facial triplets and learning a a meaningful representations that ensure the distance constraints between anchor-positive and anchor-negative self-supervisedly.

To further capture the long-range temporal dynamics within a facial sequence, we use a fixed anchor with sliding positive and negative faces to further perceive the temporal AUs' coherence and evolution characteristics.  
Fig.~\ref{fig:frame_sequence} illustrates how we construct the triplets given a facial sequence.
As can be seen, the sibling frames (e.g., $\mathbf{f}_1$ and $\mathbf{f}_2$) always show similar facial AUs with subtle difference,  while the facial frames with more temporal intervals (e.g., $\mathbf{f}_1$ and $\mathbf{f}_6$) illustrate obviously different facial AUs. Thus, we use $\mathbf{f}_{1}$ as the fixed anchor frame and the subsequent sibling pairs (the preceding frame as the positive sample and the subsequent frame as the negative) to construct a set of triplets: $((\mathbf{f}_{1}^a, \mathbf{f}_{2}^p, \mathbf{f}_{3}^n), (\mathbf{f}_{1}^a, \mathbf{f}_{3}^p, \mathbf{f}_{4}^n), \cdots, (\mathbf{f}_{1}^a, \mathbf{f}_{\text{J}-1}^p, \mathbf{f}_{\text{J}}^n))$. For each triplet, we aim to ensure that the \textit{anchor} face is closer to the positive face than it is to the negative one in the latent normalized embedding space. 

This intra-video temporal contrastive learning (TCL) paradigm naturally learns to classify the frames based on their temporal distances from an anchor frame, thus the proximity, as well as the distinctiveness between the faces within a video clip can be perceived self-supervisedly.
Our TCL has two core differences with \cite{lu2020self}: (1) We propose a weighted triplet loss. The weight can be used to suppress the ambiguous triplets and facilitates CLP better capturing the temporal dynamics. (2) We exploit the reversed counterpart of a original facial sequence to further perceive the temporal AU properties. The ambiguous triplets in the original facial sequence will be mitigated in the reversed sequence. We describe them as below.

Firstly, we observe that with the increasing intervals between $\mathbf{f}^a$ and $\mathbf{f}^p$ in the triplets, the positive sample $\mathbf{f}^p$ becomes more similar with $\mathbf{f}^n$ but less similar with $\mathbf{f}^a$. 
Optimizing such triplets might violates the temporal coherence. To mitigate this issue, we propose a weighted triplet loss $\mathcal{L}_{triplet}$ for intra-video contrastive learning,
\begin{align}
	\label{eq:triplet_loss}
	\mathcal{L}_{triplet} = \sum_{i=1}^{\text{I}} \lambda_{j} \sum_{j=2}^{\text{J}-1} \left[ D(\mathbf{f}^a_{1}, \mathbf{f}^p_{j}) -  D(\mathbf{f}^a_{1}, \mathbf{f}^n_{j+1}) + m \right]_{+},
\end{align}
where $D(\mathbf{u},\mathbf{v})$ means the Euclidean distance between sample $\mathbf{u}$ and $\mathbf{v}$. Besides, 
$\lambda_j$ denotes the weight for the $j$-th triplet, it should be monotonically descending with the increase of the temporal distances between the anchor and the positive faces. In this manner, the weight can be used to suppress the ambiguous triplets and facilitates CLP better capturing the temporal dynamics of the facial AUs.

Secondly, we further exploit the reversed counterpart of a facial sequence to further perceive the temporal AU properties. Typically, we construct another set of triplets with the reversed facial sequence, i.e.,  adopting the $\text{J}$-th facial frame as the fixed anchor face and constructing the triplets within the reversed facial sequence accordingly. Thus, we obtain $\mathcal{L}_{tcl} = \mathcal{L}_{triplet} + \mathcal{L}_{triplet}^{rev}$ to learn the frame-wisely discriminative AU representations. The proposed TCL mechanism in CLP ranks the intra-video facial frames based on their temporal distance to the anchors and encodes the temporal characteristics into the learned AU representations. More importantly, it will avoid the unexpected mode collapse that a sequence of frames collapsed to a single point. 

While TCL gives us per-frame AU representation, it has no notion of how similar are the facial AUs between different subjects. Intuitively, the representations of different subjects should be close if they have analogous facial AUs. To further learn the person-independent AU representations, we introduce a cross-identity reconstruction paradigm that extends the intra-video contrastive learning to inter-video invariance learning, which will be described in the next section.

\subsection{Cross-identity Reconstruction in CLP}
\label{sec:Cross_video_Cycle_Consistency}
To attenuate the person-specific effects in the encoded AU representations, we propose a cross-identity reconstruction (CIR) paradigm. For the AU representation vector from a subject $S$,  we exploit a dictionary of AU representations that comes from a number of different identities $C$ to reconstruct the representations of $I$. As $I$ does not exist in $C$, the AU representations from different identities will be pushed close if they show consistent or similar AUs during the reconstruction process. Thus, the person-specific nuisances would be mitigated and the generalization ability of the proposed CLP would be improved accordingly.

The pipeline of CIR is illustrated in Fig.~\ref{fig:framework_overview}, where we generalize the contrastive learning with positive pairs across videos from different identities.  
Let us denote a randomly sampled facial frame as $\mathbf{f}_{r}$, whose AU representation $\mathbf{q}_{r}$ can be encoded via the encoder in Fig.~\ref{fig:framework_overview}. Subsequently, we aim to reconstruct the nearest neighbour of $\mathbf{q}_{r}$  in a  dictionary $\mathbf{C}$. Accordingly, the reconstructed result of $\mathbf{q}_{r}$ is denoted as $\widehat{\mathbf{q}}_{r}$ .

\begin{figure}[htb]
	\centering
	\includegraphics[width=0.5\linewidth]{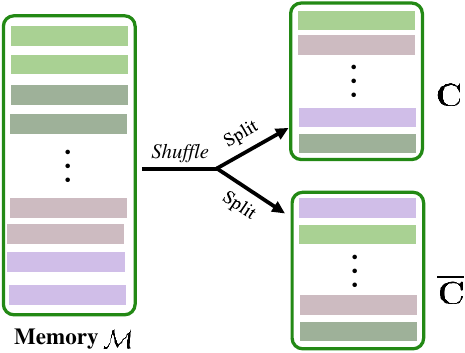}
	\caption{
		Illustration of the relationship between $\mathbf{C}$ and $\overline{\mathbf{C}}$.
	}
	\label{fig:visualize_c}
\end{figure}

To obtain the dictionary $\mathbf{C}$ that consists of a large amount of diverse AU representations, two strategies are incorporated: (1) A memory bank $\mathcal{M}$ is used to track the AU representations obtained from the preceding training batches; (2) The representations in $\mathcal{M}$ are randomly sampled from each unlabeled facial video to make the stored elements as diverse as possible. For CIR at each training iteration, we randomly split $\mathcal{M}$ into two parts: $\mathcal{M} = \{\mathbf{C}, \overline{\mathbf{C}}\}$, where $\mathbf{C}$ means the AU representation dictionary. $\overline{\mathbf{C}}$ means the left elements in $\mathcal{M}$ which will be considered as the negative instances for CIR later.  We visualize the relation between $\mathbf{C}$ and $\overline{\mathbf{C}}$ in Fig.~\ref{fig:visualize_c}.
Typically, the cosine similarity scores between $\mathbf{q}_{r}$ and each element in $\mathbf{C}$ are normalized and used as the linear combination coefficients to reconstruct $\mathbf{q}_{r}$. Formally, the coefficients $\alpha_{c}$ can be formulated as:
\begin{align}
	\label{eq:attention_score}
	\alpha_{c}  =   \frac{exp(sim(\mathbf{q}_{r}, \mathbf{C}_c)/\tau)}{\sum_{c=1}^{|\mathbf{C}|} exp(sim(\mathbf{q}_{r}, \mathbf{C}_c)/\tau)},
\end{align}
where $sim(\mathbf{u},\mathbf{v}) = \mathbf{u}^T\mathbf{v}/\|\mathbf{u}\|\|\mathbf{v}\|$ is the cosine distance measurement between the two compared vectors $\mathbf{u}$ and $\mathbf{v}$. The coefficients are normalized with \textit{softmax} operation. $\tau$ is the temperature parameter that is commonly used in the self-supervised contrastive learning methods. Accordingly, the reconstructed AU representation $\widehat{\mathbf{q}}_{r}$ can be expressed as,
\begin{align}
	\label{eq:soft_nearest_neighbor}
	\widehat{q}_{r}  =   \sum_{1 \leq c \leq |\mathbf{C}|} \alpha_{c} \mathbf{C}_{c},
\end{align}
where $|\mathbf{C}|$ is the number of the elements in the candidate neighbour set $\mathbf{C}$. With the reconstructed result $\widehat{q}_{r}$, we aim to match $\widehat{q}_{r}$  in the union collection $\widehat{\mathbf{C}} = \{\overline{\mathbf{C}}, \mathbf{k}_{r}\}$. As exemplified in Fig.~\ref{fig:framework_overview}, $\mathbf{k}_{r}$ corresponds to the representation of a pixel-level augmentation version of $\mathbf{f}_{r}$. Obviously, $\mathbf{k}_{r}$ should be the positive instance of $\widehat{\mathbf{q}}_{r}$, while the elements in $\overline{\mathbf{C}}$ should be the negative instances in the contrastive learning paradigm.
With the above intuition, CIR can be approximated if the model can distinguish the positive from the negative instances. 
Accordingly, the AU representations would be progressively cross-identity consistent during the training process without explicitly generating pseudo labels.
As to achieve this, we consider $(\widehat{\mathbf{q}}_{r}, \mathbf{k}_{r})$ as a positive pair, and $(\widehat{\mathbf{q}}_{r}, \mathbf{C}_{c})$ as numerous negative pairs for CIR. The optimization goal for CIR can thus be formulated as:
\begin{align}
	\label{eq:cvc_loss}
	\mathcal{L}_{cir} =  -\text{log} \frac{exp(sim(\widehat{\mathbf{q}}_{r}, \mathbf{k}_{r}) / \tau)}{exp(sim(\widehat{\mathbf{q}}_{r}, \mathbf{k}_{r})/ \tau +\sum_{c=1}^{|\overline{\mathbf{C}}|} exp(sim(\mathbf{q}_{r}, \overline{\mathbf{C}}_c) / \tau)}.
\end{align}

In this manner, the encoder will be optimized to perceive the AUs similarity across identities where faces with similar AUs are pulled to be close. As shown in Fig.~\ref{fig:framework_overview}, we leverage a \textit{momentum encoder} to obtain the elements in the memory bank $\mathcal{M}$. The memory is built as a queue like that in MoCo \cite{he2020momentum}, with the current mini-batch enqueued and the oldest mini-batch dequeued. It is worth noting that $\mathbf{k}_{r}$ is also encoded by the momentum encoder, thus $\mathbf{k}_{r}$ can be contrasted with $\widehat{\mathbf{q}}_{r}$ which is a linear combination of the elements in $\mathbf{C}$. 

The Encoder in CLP consists of a ResNet-34~\cite{he2016deep} backbone and two projection heads.
The backbone encodes a $512$-dimensional representations for each input facial image.
The two heads are used for \textit{TCL} and \textit{CIR}, respectively. Each head maps the $512$-dimensional representations via two fully-connected layers: $512\rightarrow256\rightarrow128$.
The  Momentum encoder consists of a ResNet-34 backbone and a CIR head.
The parameters in the momentum encoder are not shared with the encoder and are updated in a \textit{momentum} manner~\cite{he2020momentum}. Let  $w_q$ and $w_k$ denote the backbone and CIR head parameters in the \textit{encoder} and the \textit{momentum encoder}, we update $w_k$ by:
\begin{align}
	w_k^t \leftarrow \rho w_k^{t-1} + (1 - \rho)w_q^{t-1}.
\end{align}
Here $\rho$ means a momentum coefficient and was set as 0.999 in our experiments.


\begin{table*}[htb]
	\caption{F1 scores on BP4D dataset. The \textbf{bolds} denote the best in the supervised and self-supervised methods, respectively. * means the values are reported in the original papers.}
	\centering
	\label{tab:bp4d_cross_dataset}
	\begin{tabular}{|l|c|c|c|c|c|c|c|c|c|c|c|c|c|c|}
		\hline
		Methods/AU & 1 & 2 & 4 & 6 & 7 & 10 & 12 & 14 & 15 & 17 & 23 & 24 & Average \\
		\hline
		\hline
		ROI \cite{chu2017learning}*  & 36.2 & 31.6 & 43.4 & 77.1 & 73.7 & 85.0 & 87.0 & 62.6 & 45.7 & 58.0 & 38.3 & 37.4 & 56.4\\
		EAC-Net\cite{li2017eac}* &  39.0 & 35.2 & 48.6 & 76.1 & 72.9 & 81.9 & 86.2 & 58.8 & 37.5 & 59.1 & 35.9 & 35.8 & 55.9 \\
		ATF \cite{zhang2018identity} * &  39.2 & 35.2 & 45.9 & 71.6 & 71.9 & 79.0 & 83.7 & 65.5 & 33.8 & 60.0 & 37.3 & 41.8 & 55.4 \\
		IdenNet \cite{tu2019idennet}* &  50.5 & 35.9 & 50.6 & 77.2 & 74.2 & 82.9 & 85.1 & 65.3 & 42.2 & 60.8 & 42.1 & 46.5 & 59.3 \\
		DSIN \cite{corneanu2018deep}* &  51.7 & 40.4 & 56.0 & 76.1 & 73.5 & 79.9 & 85.4 & 62.7 & 37.3 & 62.9 & 38.8 & 41.6 & 58.9 \\
		JAA-Net\cite{shao2018deep}*  & 47.2 & 44.0 & 54.9 & 77.5 & 74.6 & 84.0 & 86.9 & 61.9 & 43.6 & 60.3 & 42.7 & 41.9 & 60.0 \\
		AU-RCNN \cite{ma2019r} *  & 50.2 & 43.7 & 57.0 & 78.5 & 78.5 & 82.6 & 87.0 & \textbf{67.7} & 49.1 & 62.4 & \textbf{50.4} & 49.3 & 63.0\\
		ARL \cite{shao2019facial} * & 45.8 & 39.8 & 55.1 & 75.7 & 77.2 & 82.3 & 86.6 & 58.8 & 47.6 & 62.1 & 47.4 & 55.4 & 61.1 \\
		SRERL \cite{li2019semantic}* & 46.9 & 45.3 & 55.6 & 77.1 & 78.4 & 83.5 & 87.6 & 63.9 & 52.2 & \textbf{63.9} & 47.1 & 53.3 & 62.1 \\
UGN \cite{song2021uncertain} * & 54.2 & 46.4 & 56.8 & 76.2 & 76.7 & 82.4 & 86.1 & 64.7 & 51.2 & 63.1 & 48.5 & 53.6 & 63.3 \\		
		SEV-Net \cite{yang2021exploiting}*  &  \textbf{58.2} & \textbf{50.4} & 58.3 & \textbf{81.9} & 73.9 & \textbf{87.8} & 87.5 & 61.6 & 52.6 & 62.2 & 44.6 & 47.6 & 63.9 \\
		HMP-PS \cite{song2021hybrid}*  &  53.1 & 46.1 & 56.0 & 76.5 & 76.9 & 82.1 & 86.4 & 64.8 & 51.5 & 63.0 & 49.9 & 54.5 & 63.4 \\
		Jacob \textit{et al.} \cite{jacob2021facial}* & 51.7 & 49.3 & \textbf{61.0} & 77.8 & \textbf{79.5} & 82.9 & 86.3 & 67.6 & 51.9 & 63.0 & 43.7 & \textbf{56.3} & \textbf{64.2} \\
		SO-Net \cite{yang2020set}* & 40.2 & 46.2 & 56.0 & 79.3 & 73.5 & 84.2 & \textbf{90.8} & 64.7 & \textbf{55.9} & 61.0 & 37.4 & 40.2 & 60.8 \\
		PIAP \cite{tang2021piap}* & 54.2 & 47.1 & 54.0 & 79.0 & 78.2 & 86.3 & 89.5 & 66.1 & 49.7 & 63.2 & 49.4 & 52.0 & 64.1\\
		ResNet-34 & 46.7 & 44.0 & 49.7 & 73.0 & 74.5 & 81.4 & 85.5 & 61.6 & 48.2 & 57.4 & 44.3 & 44.5 & 59.2 \\
		\hline
		\hline
		SimCLR \cite{chen2020simple}  &  38.0 & 36.4 & 37.2 & 66.6 & 64.7 & 76.2 & 76.2 & 51.1 & 29.8 & 56.1 & 27.5 & 37.7 & 49.8 \\
		MoCo \cite{he2020momentum} & 30.8 & 41.3 & 42.1 & 70.2 & 70.4 & 78.7 & 82.5 & 53.3 & 25.2 & 59.1 & 31.5 & 34.3 & 51.6  \\
		SimSiam \cite{chen2021exploring} & 38.6 & 39.0 & 37.2 & 68.2 & 65.2 & 76.1 & 77.6 & 56.3 & 37.3 & 57.9 & 25.4 & 39.7 & 51.5 \\
		CVC \cite{wu2021contrastive} & 43.9 & 47.8 & 38.7 & 67.0 & 70.4 & 81.8 & 84.4 & 57.5 & 39.5 & 49.3 & 27.1 & 43.6 & 54.2 \\
		\hline
		Lu \textit{et al.} \cite{lu2020self}* &  42.3 & 24.3 & 44.1 & 71.8 & 67.8 & 77.6 & 83.3 & 61.2 & 31.6 & 51.6 & 29.8 & 38.6 & 52.0 \\
		Fab-Net\cite{wiles2018self}* &  43.3 & 35.7 & 41.6 & 72.9 & 63.0 & 75.9 & 83.5 & 57.7 & 26.5 & 48.2 & 33.6 & 42.4 & 52.0 \\
		TAE \cite{li2020learning}* &  47.0 & 45.9 & \textbf{50.9} & 74.7 & 72.0 & \textbf{82.4} & \textbf{85.6} & 62.3 & 48.1 & 62.3 & \textbf{45.9} & 46.3 & 60.3 \\
		EmoCo \cite{sun2021emotion}* & 45.4 & 30.5 & 55.5 & 76.1 & 75.7 & 84.4 & 87.6 & 66.6 & 39.6 & 59.1 & 41.3 & 49.8 & 59.3 \\
		\textbf{CLP (Ours)}  &  \textbf{47.7} & \textbf{50.9} & 49.5 & \textbf{75.8} & \textbf{78.7} & 80.2 & 84.1 & \textbf{67.1} & \textbf{52.0} & \textbf{62.7} & 45.7 & \textbf{54.8} & \textbf{62.4} \\
		
		\hline
	\end{tabular}
\end{table*}

\begin{table*}[htb]
	\centering
	\caption{F1 scores on DISFA dataset.  The \textbf{bolds} denote the best in the supervised and self-supervised methods, respectively. * means the values are reported in the original papers.}
	\label{tab:disfa_cross_dataset}
	\begin{tabular}{|l|c|c|c|c|c|c|c|c|c|}
		\hline
		Methods/AU & 1 & 2 & 4 & 6 & 9 & 12 & 25 & 26 & Average \\
		\hline
		\hline
		ROI \cite{li2017eac}*  &  41.5 & 26.4 & 66.4 & 50.7 & \textbf{80.5} & \textbf{89.3} & 88.9 & 15.6 & 48.5 \\
		EAC-Net \cite{li2017eac}*  & 41.5 & 26.4 & 66.4 & 50.7 & 80.5 & 89.3 & 88.9 & 15.6 & 48.5 \\
		ATF \cite{zhang2018identity}*  & 45.2 & 39.7 & 47.1 & 48.6 & 32.0 & 55.0 & 86.4 & 39.2 & 49.2 \\
		IdenNet \cite{tu2019idennet}* & 25.5 & 34.8 & 64.5 & 45.2 & 44.6 & 70.7 & 81.0 & 55.0 & 52.6 \\ 
		DSIN \cite{corneanu2018deep}*  & 42.4 & 39.0 & 68.4 & 28.6 & 46.8 & 70.8 &  90.4 & 42.2 & 53.6 \\
		JAA-Net \cite{shao2018deep}*  & 43.7 & 46.2 & 56.0 & 41.4 & 44.7 & 69.6 & 88.3 & 58.4 & 56.0 \\
		AU-RCNN \cite{ma2019r} *  & 32.1 & 25.9 & 59.8 & 55.3 & 39.8 & 67.7 & 77.4 & 52.6 & 51.3 \\
		ARL \cite{shao2019facial} * & 43.9 & 42.1 & 63.6 & 41.8 & 40.0 & 76.2 & 95.2 & 66.8 & 58.7 \\ 
		SRERL \cite{li2019semantic}* & 45.7 & 47.8 & 59.6 & 47.1 & 45.6 & 73.5 & 84.3 & 43.6 & 55.9 \\
		UGN \cite{song2021uncertain} * & 43.3 & 48.1 & 63.4 & 49.5 & 48.2 & 72.9 & 90.8 & 59.0 & 60.0 \\
		SEV-Net\cite{yang2021exploiting}*  &  55.3 & \textbf{53.1} & 61.5 & 53.6 & 38.2 & 71.6 & \textbf{95.7} & 41.5 & 58.8 \\
		HMP-PS \cite{song2021hybrid}*  & 38.0 & 45.9 & 65.2 & 50.9 & 50.8 & 76.0 & 93.3 & \textbf{67.6} & 61.0 \\
		Jacob \textit{et al.} \cite{jacob2021facial}* & 46.1 & 48.6 & \textbf{72.8} & 56.7 & 50.0 & 72.1 & 90.8 & 55.4 & 61.5 \\
		SO-Net \cite{yang2020set}* &  33.8 & 44.5 & 70.3 & \textbf{57.6} & 39.7 & 78.2 & 86.7 & 57.3 & 58.5 \\
		PIAP \cite{tang2021piap}* & \textbf{50.2} & 51.8 & 71.9 & 50.6 & 54.5 & 79.7 & 94.1 & 57.2 & \textbf{63.8} \\
		ResNet-34 &  38.0 & 33.1 & 51.8 & 46.2 & 34.2 & 65.4 & 85.4 & 56.9 & 51.3 \\
		\hline
		\hline 
		SimCLR \cite{chen2020simple} & 21.2 & 23.3 & 47.5 & 42.4 & 35.5 & 66.8 & 81.5 & 52.7 & 46.4 \\
		MoCo \cite{he2020momentum} & 22.7 & 18.2 & 45.9 & 45.4 & 34.1 & 72.9 & 83.4 & 54.5 & 47.1 \\
		SimSiam \cite{chen2021exploring} & 35.5 & 25.5 & 58.1 & 53.8 & 32.4 & 74.4 & 79.0 & 55.7 & 51.8 \\
		CVC \cite{wu2021contrastive} & 30.3 & 20.9 & 56.4 & 49.5 & 26.3 & \textbf{75.5} & 79.1 & 51.8 & 48.6 \\ 
		\hline
		Lu \textit{et al.} \cite{lu2020self}* &  18.7 & 27.4 & 35.1 & 33.6 & 20.7 & 67.5 & 68.0 & 43.8  & 39.4 \\
		Fab-Net \cite{wiles2018self}* &  15.5 & 16.2 & 43.2 & 50.4 & 23.2 & 69.6 & 72.4 & 42.4  & 41.6 \\
		TAE \cite{li2020learning}* &  21.4 & 19.6 & \textbf{64.5} & 46.8 & \textbf{44.0} & 73.2 & \textbf{85.1} & 55.3 & 51.5 \\
		EmoCo \cite{sun2021emotion}* &  34.3 & 31.9 & 63.9 & 52.5 & 44.0 & 77.0 & 78.3 & 44.2 & 53.3 \\
		\textbf{CLP (Ours)}  &  \textbf{42.4} & \textbf{38.7} & 63.5 & \textbf{59.7} & 38.9 & 73.0 & 85.0 & \textbf{58.1} & \textbf{57.4} \\
		\hline
	\end{tabular}
\end{table*}

\begin{table*}[htb]
	\centering
	\caption{F1 score on GFT dataset.  The \textbf{bolds} denote the best in the supervised and self-supervised methods, respectively. * means the values are reported in the original papers.}
	\label{tab:gft_cross_dataset}
	\begin{tabular}{|l|c|c|c|c|c|c|c|c|c|c|c|}
		\hline
		Methods/AU & 1 & 2 & 4 & 6 & 10 & 12 & 14 & 15 & 23 & 24 & Average \\
		\hline
		\hline  
		AlexNet \cite{girard2017sayette}*  &  44 & 46 & 2 & 73 & 72 & 82 & 5 & 19 & 43 & 42 & 42.8  \\
		MAL \cite{li2021meta} & \textbf{52.4} & \textbf{57.0} & \textbf{54.1} & \textbf{74.5} & \textbf{78.0} & \textbf{84.9} & \textbf{43.1} & \textbf{47.7} & \textbf{54.4} & \textbf{51.9} & \textbf{59.8} \\
		ResNet-34  & 43.2 & 48.2 & 25.9 & 74.0 & 73.7 & 83.5 & 37.5 & 44.0 & 51.7 & 48.4 & 53.0 \\
		\hline
		\hline 
		SimCLR \cite{chen2020simple} & 39.6 & 48.3 &  5.6 & 80.7 & 76.2 & 80.6 & 18.1 & 41.6 & 46.1 & 43.8 & 48.1 \\
		MoCo \cite{he2020momentum} & 35.9 & 45.4 & 13.5 & \textbf{83.4} & 71.3 & 78.1 & 23.3 & 37.3 & 26.6 & 50.7 & 46.5 \\
		SimSiam  \cite{chen2021exploring} & 42.3 & 48.5 & 38.1 & 77.8 & 75.7 & 75.7 & 29.6 & 37.8 & 48.0 & 44.7 & 51.8 \\
		CVC \cite{wu2021contrastive}  & 45.9 & 46.5 & 24.5 & 79.2 & 68.8 & 80.3 & 38.6 & 41.8 & 37.9 & 39.8 & 50.3 \\
		\hline
		Fab-Net\cite{wiles2018self}* &  44.4 & 42.3 & 9.4 & 60.6 & 68.7 & 70.4 & 8.7 & 1.7 & 5.5 & 20.8 & 33.3 \\
		TAE \cite{li2020learning}* &  \textbf{46.3} & 48.8 & 13.4 & 76.7 & 74.8 & 81.8 & 19.9 & 42.3 & 50.6 & \textbf{50.0} & 50.5 \\
		EmoCo \cite{sun2021emotion}* & 51.8 & 42.9 & 22.9 & 79.8 & 77.0 & 85.2 & \textbf{23.4} & 42.5 & 55.4 & 49.6 & 53.0 \\ 
		\textbf{CLP (Ours)} & 44.6 & \textbf{58.7} &
		\textbf{34.7} & 75.9 & \textbf{78.6} & \textbf{86.6} & 20.3 & \textbf{44.8} & \textbf{56.4} & 42.2 & \textbf{54.3} \\ 
		\hline
	\end{tabular}
\end{table*}

\subsection{Training objective}
We linearly combine the temporal contrastive and CIR objectives to for training. The full loss is formulated as,
\begin{align}
	\label{eq:total_loss}
	\mathcal{L}_{tot} =  \mathcal{L}_{tcl}  + \beta \mathcal{L}_{cir}.
\end{align}
The hyper-paramter $\beta$ controls the importance of $\mathcal{L}_{cir}$ and will be discussed in Sec.~\ref{sec:experimental_details}.

\section{Experiment}
In this section, we first present the implementation details of CLP. Then we validate the effectiveness of CLP on three popular AU datasets, including BP4D\cite{zhang2013high},  DISFA\cite{mavadati2013disfa} and GFT \cite{girard2017sayette}.
We compare CLP with other self-supervised AU detection methods, representative contrastive learning methods, and state-of-the-art AU detection methods on the three AU datasets. Finally, we analyze the effectiveness of each component in CLP via ablation studies.

\subsection{Implementation details}
\label{sec:experimental_details}

\textbf{Training of CLP:} 
Our proposed CLP was trained on the VoxCeleb2 dataset \cite{chung2018voxceleb2}. The facial videos in VoxCeleb2 dataset consists of around 6,000 subjects. For each subject in VoxCeleb2, we randomly selected 8 videos and extracted the facial frames at 25 frames per second. The extracted facial frames were detected and aligned via an open-source face detector SeetaFace \footnote{https://github.com/seetaface/SeetaFaceEngine}. To avoid discontinuities, we discarded the sequences where a face is not detected.  Finally, there are approximately 8.63M unlabeled facial frames for training.  We applied a set of image augmentations during training, including uniform rotation ($\pm$10 degree), random horizontal flip, random color jitter, and random cropping.

For \textit{TCL}, we randomly selected a consecutive sequence $F_i$ with 9 facial frames. 
According to our experiments, if we set $s = 1$, the adjacent frames will be too similar and the temporal contrastive learning loss in CLP would be hard to optimize. When we set $s \geq 3$, the temporal evolution characteristics will be reduced and the CLP shows degraded AU detection performance. Finally, we use $s = 2$ in our experiments.
We set the length of a consecutive facial sequence as $|F_i| = 9$ according to the AU detection performance on the BP4D dataset. Reducing the length of a facial sequence will show degraded AU detection performance (we obtain 59.7 average F1 score with $|F_i| = 7$). It indicates a facial sequence should contains enough faces to facilitate encoding the long-range temporal dynamics.
When we enlarge the length of a facial sequence with $|F_i| = 11$, we obtain similar AU detection performance. Further increasing the length of a facial sequence induces consistent degradation (we obtain 60.3/58.7 average F1 score with $|F_i| = 13/15$).  It might be explained that longer faical sequence will introduce more triplet ambiguity.
The margin was set as $m = 0.03$ in Equ.~\ref{eq:triplet_loss} via manual grid search.

For \textit{CIR}, we randomly selected $L=2$ discontinuous frames per-video. The size of $\mathcal{M}$ was set as 65536 and we used half of the elements for the candidate neighbour set $\mathbf{C}$ and the left half for $\overline{\mathbf{C}}$. The loss weight $\beta$ in Equ.~\ref{eq:total_loss} was set as $0.1$ via grid search. We used ResNet-34 \cite{he2016deep} as the backbone for the encoder and the momentum encoder.  According to \cite{he2016deep}, when the input size is set as $224 \times 224$, the ResNet-34 model involves 3.6 GFLOPs (FLOPs means floating point operations) per inference. 
The model was implemented in PyTorch \cite{paszke2017automatic} using 4 GPUs, each having a batch size of 16 sequences. We set the initial learning rate of 0.001. It took around 200 epochs and 2 days to converge.

\textbf{Evaluation protocols:} We discarded the projection heads and mapped the AU representation with a linear AU classifier. A batch-norm layer  followed by a linear fully connected layer with no bias constitute the linear classifier. The linear classifier was trained with a binary cross-entropy loss for each AU. The parameters in the encoder were frozen except that in the batch-norm layers \cite{he2020momentum}.  Since the activated AUs in the AU datasets are widely skewed, we reweighted the samples from under-represented categories inversely proportional to the AU class frequencies.
We used the F1 score ($F1=\frac{2RP}{R+P}$) to quantitatively evaluate the performance of the method, where $R$ and $P$ denote recall and precision, respectively.
In addition, we computed the average performance over all AUs. The overall performance of CLP and the compared methods are measured by the average F1 score. We showed the AU detection results as $F1 \times 100$. For evaluation, the ResNet-34 model in our proposed CLP consists of 21.8M parameters and involves 3.68 GFLOPs (FLOPs means floating point operations) per inference~\cite{he2016deep}. According to our experiments, when the batch size is set as 16, our proposed CLP takes 15.6 ms for a single model inference with a NVIDIA TITAN Xp GPU (12 GB memory). While for CPU-based inference (Intel Core i7-12700KF), our proposed CLP takes 248.6 ms for one model inference. 

\subsubsection{Datasets} For AU detection, We adopted BP4D \cite{zhang2013high}, DISFA \cite{mavadati2013disfa}, and GFT \cite{girard2017sayette} datasets. 
Among the three datasets, BP4D is a spontaneous FACS dataset that consists of 328 videos for 41 patients (18 males and 23 females). Each subject participates in eight sessions, with both 2D and 3D movies capturing their spontaneous facial action units. There are around 142,000 frames with AU annotations of occurrence or absence for the 328 videos.
 There are 27 individuals in DISFA, including 12 females and 15 males. To elicit facial AUs, each subject is invited to watch a 4-minute film. The strengths of the AUs range from 0 to 5. We acquired over 130,000 AU-annotated images in the DISFA dataset in our tests by classifying images with intensities greater than 1 as active.
GFT contains 96 subjects and the subjects are arranged in 32 three-person groups. Moderate out-of-plane head motion and occlusion are presented in the video frames, making AU detection in the GFT dataset challenging. 
We performed a 3-fold cross-validation on the BP4D and DISFA datasets by splitting the images into three folds in a subject-independent manner. For evaluation, we used 12 AUs from the BP4D dataset and 8 AUs from the DISFA dataset. For GFT dataset, We used total ten AUs to evaluate the AU representations on the GFT dataset. We acquired roughly 108000 facial photos for training and 24600 images following \cite{girard2017sayette} for evaluation using the original train/test splits.

To verify the generalization of the feature, we evaluate our proposed CLP on RAF-DB~\cite{li2017reliable} and AffectNet~\cite{mollahosseini2017affectnet} datasets to see how well the representations can generalize to facial expression recognition (FER). RAF-DB dataset contains approximately 30,000 facial images annotated with basic or compound expressions by 40 trained human coders. In our experiment, only images with these seven basic emotions were used, including 12,271 images as training data and 3,068 images as test data. AffectNet is the largest dataset with annotated facial emotions. It contains about 400,000 images manually annotated for the presence of seven discrete facial expressions and the intensity of valence and arousal. We only used the images with neutral and 6 basic emotions, containing 280,000 training samples and 3,500 test samples.

\subsection{Comparison with the state-of-the-art methods}
\label{sec:Comparison_with_the state-of-the-art}

Tab.~\ref{tab:bp4d_cross_dataset}, Tab.~\ref{tab:disfa_cross_dataset}, Tab.~\ref{tab:gft_cross_dataset} report the F1-scores of CLP and the compared methods on BP4D\cite{zhang2013high},  DISFA\cite{mavadati2013disfa} and GFT \cite{girard2017sayette} datasets.
To investigate the distinctiveness of the AU representations in CLP, we first compare CLP with the representative supervised state-of-the-art AU detection methods. Then we compare CLP with the general self-supervised contrastive representation learning methods, as well as the typical self-supervised AU detection methods.
On each dataset, we trained a 34-layer ResNet \cite{he2016deep} pre-trained by ImageNet \cite{deng2009imagenet} for quantitative and qualitative comparisons.

\begin{table}[t]
	\centering
	\caption{Ablation experiments on BP4D and DISFA datasets.}
	\label{tab:ablationstudy}
	\begin{tabular}{c|cccccc}
		\hline
		Dataset & $\mathcal{L}_{tcl}$ & $\mathcal{L}_{cir}$ & $\lambda_j$  & $|C|$ & F1 \\
		\hline
		 BP4D & $\checkmark$ & $\times$ & $1$ &  $\times$ & 52.0 \\
		 BP4D & $\checkmark$ & $\times$ & $1 / \sqrt{j}$ &  $\times$ & 55.8 \\
		 BP4D & $\times$ & $\checkmark$ & $1 / \sqrt{j}$ &  32768 & 59.3 \\
		 \hline
		 BP4D & $\checkmark$ & $\checkmark$ & $1 / \sqrt{j}$ & 32768 & \textbf{62.4} \\
		 BP4D & $\checkmark$ & $\checkmark$ & $exp(-j)$ &  32768 & 61.3 \\
		 BP4D & $\checkmark$ & $\checkmark$ & $1 / j$ &  32768 & 60.5 \\
		 \hline
		 
		 BP4D & $\checkmark$ & $\checkmark$ &  $1 / \sqrt{j}$ &  40960 & 58.3 \\
		 BP4D & $\checkmark$ & $\checkmark$ & $1 / \sqrt{j}$ &  16384 & 61.3 \\
		 BP4D & $\checkmark$ & $\checkmark$ & $1 / \sqrt{j}$ &  8192 & 59.1 \\
		 \hline
		 \hline
		 DISFA & $\checkmark$ & $\times$ & $1$ &  $\times$ & 39.4 \\
		 DISFA & $\checkmark$ & $\times$ & $1 / \sqrt{j}$ &  $\times$ & 47.2 \\
		 DISFA & $\times$ & $\checkmark$ & $1 / \sqrt{j}$ &  32768 & 54.2 \\
		 \hline
		 DISFA & $\checkmark$ & $\checkmark$ & $1 / \sqrt{j}$ & 32768 & \textbf{57.4} \\
		 DISFA & $\checkmark$ & $\checkmark$ & $exp(-j)$ &  32768 & 57.0 \\
		 DISFA & $\checkmark$ & $\checkmark$ & $1 / j$ &  32768 & 55.8 \\
		 \hline
		 DISFA & $\checkmark$ & $\checkmark$ &  $1 / \sqrt{j}$ &  40960 & 56.6 \\
		 DISFA & $\checkmark$ & $\checkmark$ & $1 / \sqrt{j}$ &  16384 & 56.7 \\
		 DISFA & $\checkmark$ & $\checkmark$ & $1 / \sqrt{j}$ &  8192 & 50.3 \\
		 \hline
	\end{tabular}
\end{table}

\begin{figure*}[htb]
	\centering
	\includegraphics[width=0.9\linewidth]{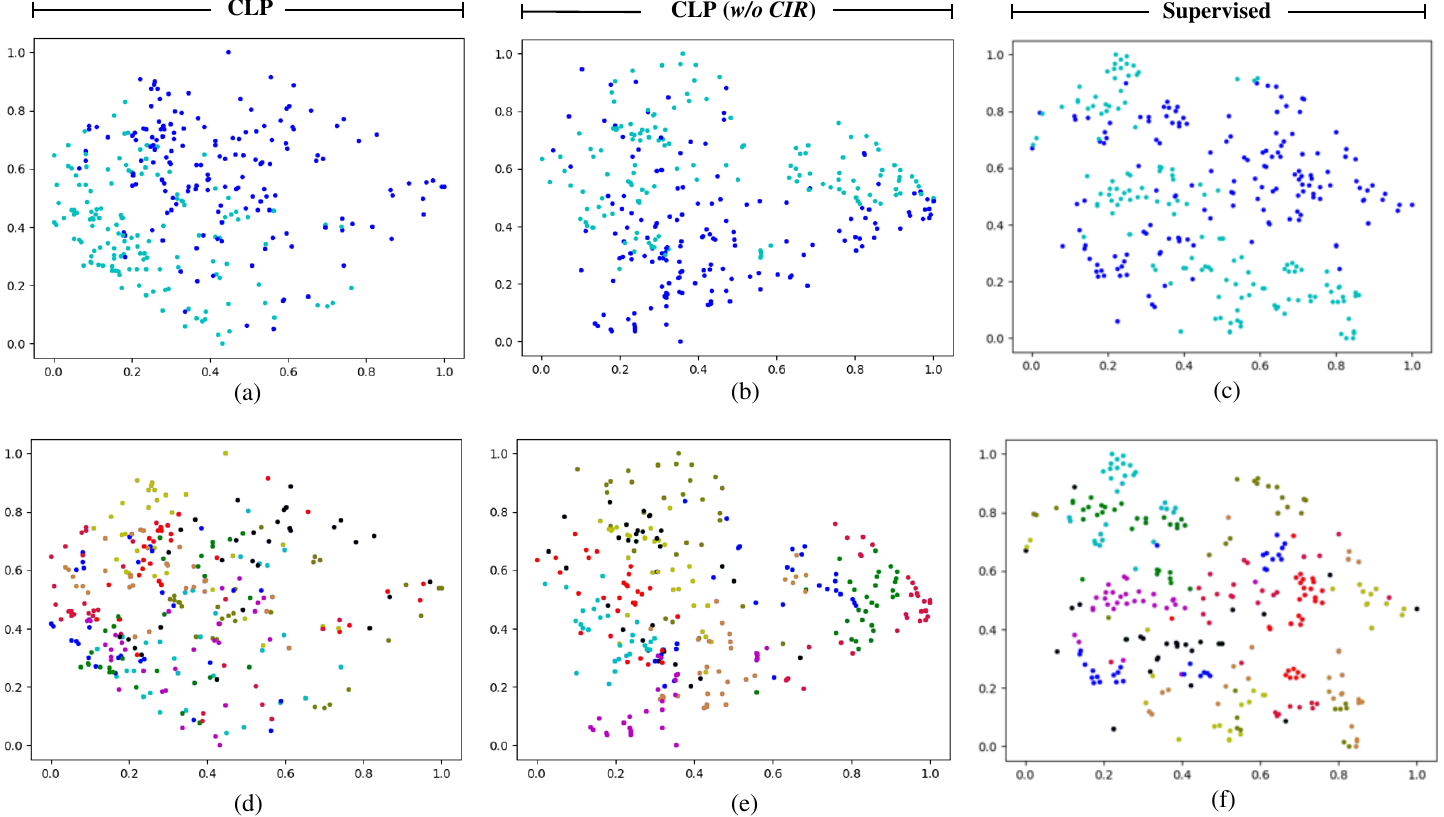}
	\caption{
		Feature visualization on BP4D dataset. 
		Top row: colors indicates whether AU12 exists. Bottom row: colors means the subjects. It is clear that CLP-learned representations are more invariant w.r.t subjects.
	}
	\label{fig:bp4d_feature_visualize}
\end{figure*}

\subsubsection{Comparison with supervised AU detection methods} \label{sec:Comparisons_with_supervised_methods}
We compare CLP with ROI \cite{li2017action}, EAC-Net \cite{li2017eac}, DSIN \cite{corneanu2018deep}, DSIN \cite{corneanu2018deep}, AU-RCNN \cite{ma2019r},  ATF \cite{zhang2018identity}, IdenNet \cite{tu2019idennet}, JAA-Net \cite{shao2018deep}, SRERL \cite{li2019semantic}, SEV-Net \cite{yang2021exploiting}, SO-Net \cite{yang2020set}, HMP-PS \cite{song2021hybrid}, MAL \cite{li2021meta} and the transformer-based \cite{devlin2018bert} AU detection method proposed by Jacob \textit{et al.} \cite{jacob2021facial}.
Among the state-of-the-art supervised AU detection methods,  ROI \cite{li2017action},  JAA-Net \cite{shao2018deep}, DSIN \cite{corneanu2018deep}, AU-RCNN \cite{ma2019r}, ARL \cite{shao2019facial}, SRERL \cite{li2019semantic} extract the regional facial features with manually selected facial landmarks and learn the region-specific AU features with exclusive CNN branches. 
Besides, PIAP \cite{tang2021piap} learns a pixel-level attention map for each AU and eliminates the person-specific features with the assistance of an auxiliary face recognition model. These methods are evaluated and compared with our proposed CLP under the same protocol on BP4D and DISFA datasets.

Tab.~\ref{tab:bp4d_cross_dataset}, Tab.~\ref{tab:disfa_cross_dataset} illustrate the AU detection performance comparison. Our proposed CLP is comparable to most of the supervised methods. Specially, It outperforms ROI \cite{li2017action}, EAC-Net \cite{li2017eac}, DSIN \cite{corneanu2018deep}, JAA-Net \cite{shao2018deep}, SRERL \cite{li2019semantic}, ATF \cite{zhang2018identity}, IdenNet \cite{tu2019idennet} on BP4D and DISFA dataset.
These supervised approaches use artificially defined multiple local facial regions to learn the region-specific AU representations, which is not the case in CLP. 
CLP is also comparable with SO-Net \cite{yang2020set} on the two AU datasets.
Notably,  CLP outperforms the supervised ResNet-34 pre-trained with ImageNet with consistent improvements on the three AU datasets (+3.2\% on BP4D, +6.1\% on DISFA, +1.3\% on GFT), indicating the superiority of the CLP-learned AU representations.

CLP lags behind SEV-Net\cite{yang2021exploiting}, HMP-PS\cite{song2021hybrid}, Jacob \textit{et al.} \cite{jacob2021facial} and PIAP \cite{tang2021piap}. On the BP4D and DISFA datasets, there are two causes for the performance disparity between our proposed CLP and a few state-of-the-art AU detection methods: 1)  CLP just uses a vanilla ResNet-34 that does not rely on facial landmarks or an attention strategy. Since AUs are reflected by the motions of specific facial muscles and are tightly correlated with the fine-grained local facial areas, learning patch-specific AU representations is useful and has been commonly used in supervised AU detection methods \cite{tang2021piap, li2017action, li2017eac, corneanu2018deep, shao2018deep, li2019semantic}. 2) The fundamental relationships between distinct AUs are not explicitly captured by CLP. As AU detection is a multi-label classification problem with some inherent relationships among AUs, explicitly capturing AU dependencies has been verified to be beneficial \cite{jacob2021facial,song2021hybrid,li2019semantic ,song2021uncertain,corneanu2018deep}. Notably, CLPs achieve competitive AU detection results with a standard ResNet-34 and require no specific neural network architecture designs.

\begin{figure*}[htb]
	\centering
	\includegraphics[width=0.9\linewidth]{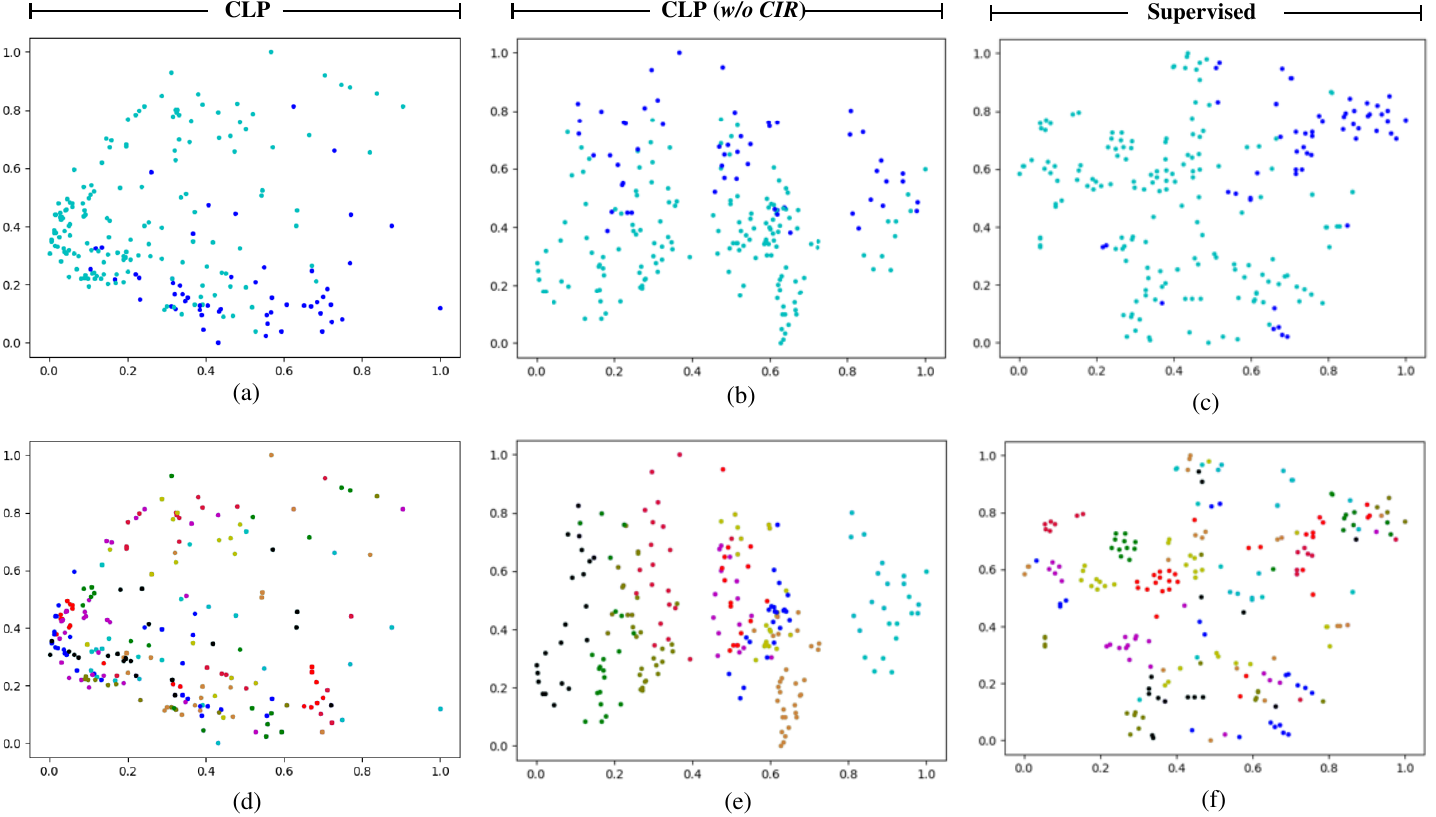}
	\caption{
		Feature visualization on DISFA dataset. 
		Top row: colors indicates whether AU12 exists. Bottom row: colors means the subjects. It is clear that CLP-learned representations are more invariant w.r.t subjects.
	}
	\label{fig:disfa_feature_visualize}
\end{figure*}

\subsubsection{Comparison with the contrastive learning methods}  
We compare CLP with the representative contrastive representation learning methods, including SimCLR \cite{chen2020simple}, MoCo \cite{he2020momentum}, SimSiam \cite{chen2021exploring} and CVC \cite{wu2021contrastive}. We re-trained the models on the same VoxCeleb2 \cite{chung2018voxceleb2} dataset using the released codes. For all these methods, we tried our best to tune their parameters to report the best results. As shown in Tab.~\ref{tab:bp4d_cross_dataset}, Tab.~\ref{tab:disfa_cross_dataset}, Tab.~\ref{tab:gft_cross_dataset}, CLP consistently outperforms them in the average F1 score (+8.2\% on BP4D, +5.6\% on DISFA, +2.5\% on GFT). It is because the general representations are not AU specific.
Although the representations in SimCLR, MoCo and SimSiam encode the intra-image invariance, they are not trained to be AU-discriminative within a short video clip.
CVC explores the cross-video relation by using cross-video cycle-consistency. However, CVC aims to learn the intra-video invariant representations, which are not frame-wisely discriminative within a short video clip.
In contrast, CLP aims to learn AU representations that are temporally discriminative to perceive the temporal coherence and evolution characteristics of the AUs. In addition, CLP-learned AU representations are tasked to be consistent across different identities that show similar AUs. Thus the representations are tasked to perceive the high-level AU semantics and neglect the undesired identity information.

\begin{figure*}[htb]
	\centering
	\includegraphics[width=0.9\linewidth]{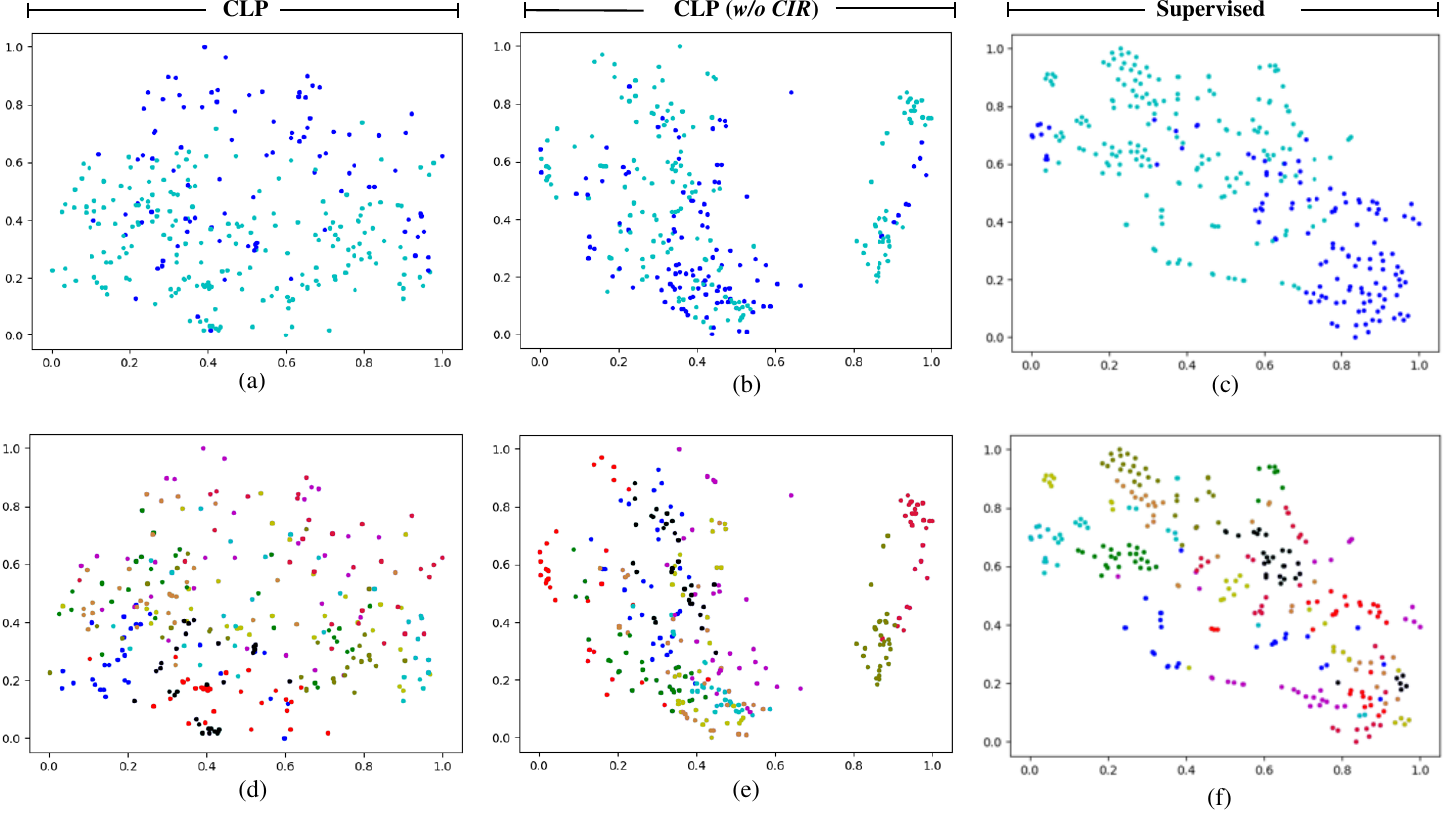}
	\caption{
		Feature visualization on GFT dataset. 
		Top row: colors indicates whether AU12 exists. Bottom row: colors means the subjects. It is clear that CLP-learned representations are more invariant w.r.t subjects.
	}
	\label{fig:gft_feature_visualize}
\end{figure*}

\begin{table*}[htb]
	\centering
	\caption{Comparison on RAF-DB dataset. \textbf{Bold} and \underline{underline} denote the best among the self-supervised and supervised methods, respectively.}
	\label{tab:rafdb_tae_test}
	\begin{tabular}{|c|c|c|c|c|c|c|c|c|}
		\hline
		Methods & Neutral & Anger & Disgust & Fear & Happy & Sad & Surprise & Average \\
		\hline
		\hline
		AlexNet \cite{krizhevsky2017imagenet}  & 60.2 & 58.6 & 21.9 & 39.2 & 86.2 & 60.9 & 62.3 & 55.6 \\
		VGG16 \cite{simonyan2014very}     & 59.9 & 68.5 & 27.5 & 35.1 & 85.3 & 64.9 & 66.3 & 58.2 \\
		gACNN \cite{li2018occlusion}   & 84.3 & 78.4 & 53.1 & 55.4 & 93.2 & 82.9 & \underline{86.3} & 76.2 \\
		VTFF \cite{9585378} & \underline{87.5} & \underline{85.8} & \underline{68.1} & \underline{64.8} & \underline{94.1} & \underline{87.2} & 85.4 & \underline{81.8} \\
		\hline
		\hline
		DeformAE \cite{shu2018deforming} &  37.1 & 53.7 & 46.9 & 48.7 & 67.1 & 29.1 & 69.9 & 50.4  \\
		Fab-Net \cite{wiles2018self} &  36.8 & 53.7 & 37.5 & 51.3 & 82.7 & 27.8 & 75.4 & 52.2  \\
		TAE \cite{li2020learning} & 62.8 & 58.0 & 45.0 & \textbf{58.1} & 76.0 & 45.8 & 64.4 & 58.6  \\
		\textbf{CLP (Ours)}  & \textbf{63.5} & \textbf{60.9} & \textbf{48.3} & 52.3 & \textbf{86.8} & \textbf{46.5} & \textbf{77.8} & \textbf{62.3}  \\
		\hline
	\end{tabular}
\end{table*}

\begin{table*}[htb]
	\centering
	\caption{Comparison on AffectNet dataset. \textbf{Bold} and \underline{underline} denote the best among the self-supervised and supervised methods, respectively.}
	\label{tab:affectNet_tae_test}
	\begin{tabular}{|c|c|c|c|c|c|c|c|c|}
		\hline
		Methods & Neutral & Anger & Disgust & Fear & Happy & Sad & Surprise & Average \\
		\hline
		\hline
		AlexNet \cite{krizhevsky2017imagenet} & $-$ & $-$ & $-$ & $-$ & $-$ & $-$ & $-$ & 47.0 \\
		VGG16 \cite{simonyan2014very}     & \underline{89.6} & 53.4 & 20.6 & 32.0 & 90.0 & 35.0 & 37.2 & 51.1 \\
		gACNN \cite{li2018occlusion}   & 73.4 & \underline{66.2} & 32.6 & 46.2 & \underline{93.8} & 55.8 & 43.4 & 58.8 \\
		VTFF \cite{9585378} & 65.0 & 61.2 & \underline{53.0} & \underline{60.4} & 88.4 & \underline{60.8} & \underline{64.8} & \underline{64.8} \\
		\hline
		\hline
		DeformAE \cite{shu2018deforming} &  33.2 & 37.2 & 35.2 & 28.4 & 79.8 & 20.2 & \textbf{62.6} & 42.4  \\
		FAB-Net \cite{wiles2018self} & 38.6 & 30.6 & \textbf{48.4} & 32.1 & 82.2 & 35.6 & 51.4 & 45.6  \\  
		TAE~\cite{li2020learning} &  44.4 & 38.6 & 46.8 & 40.4 & 78.0 & \textbf{40.8} & 54.4 & 49.1  \\
		\textbf{CLP (Ours)}  & \textbf{46.1} & \textbf{45.7} & 47.2 & \textbf{42.9} & \textbf{83.3} & 39.8 & 58.4 & \textbf{51.9}  \\
		\hline
	\end{tabular}
\end{table*}

We additionally compare CLP with the representative self-supervised AU detection methods, including Fab-Net \cite{wiles2018self}, Lu \textit{et al.} \cite{lu2020self}, TAE \cite{li2020learning}, EmoCo \cite{sun2021emotion}. 
As shown in Tab.~\ref{tab:bp4d_cross_dataset}, Tab.~\ref{tab:disfa_cross_dataset}, Tab.~\ref{tab:gft_cross_dataset},
the CLP-learned AU representation outperforms other self-supervised AU detection methods with no exception, with 2.1\%, 5.9\%, 3.8\% improvements on BP4D, DISFA, GFT dataset, respectively. Obviously, CLP outperforms the general contrastive learning methods and the self-supervised AU detection approaches on various AUs, e.g., AU1 (inner brow raiser), AU2 (outer brow raiser), AU6 (cheek raiser), AU7 (lid tightener), AU14 (dimpler), AU15 (lip corner depressor), AU17 (chin raiser), AU24 (lip pressor)on BP4D dataset. Among the compared methods,
Fab-Net and TAE are trained on pairs of unlabeled facial frames and require the per-pixel reconstruction from the source to the target facial frames. However, pixel-level generation may suffer from the illumination inconsistency between the source and target frames. Although EmoCo adopts enormous facial images annotated with six universal facial expressions for AU representation learning, it might suffer from the ambiguous relations between the in-the-wild facial expressions and the AUs \cite{yan2020raf}. In contrast, CLP directly learns the AU representations using the objective functions derived from the prior knowledge of AU. Besides, CLP generalizes the intra-video contrastive learning to inter-video invariance learning to force the AU representations to be cross-identity consistent. 

\begin{figure*}[htb]
	\centering
	\includegraphics[width=0.8\linewidth]{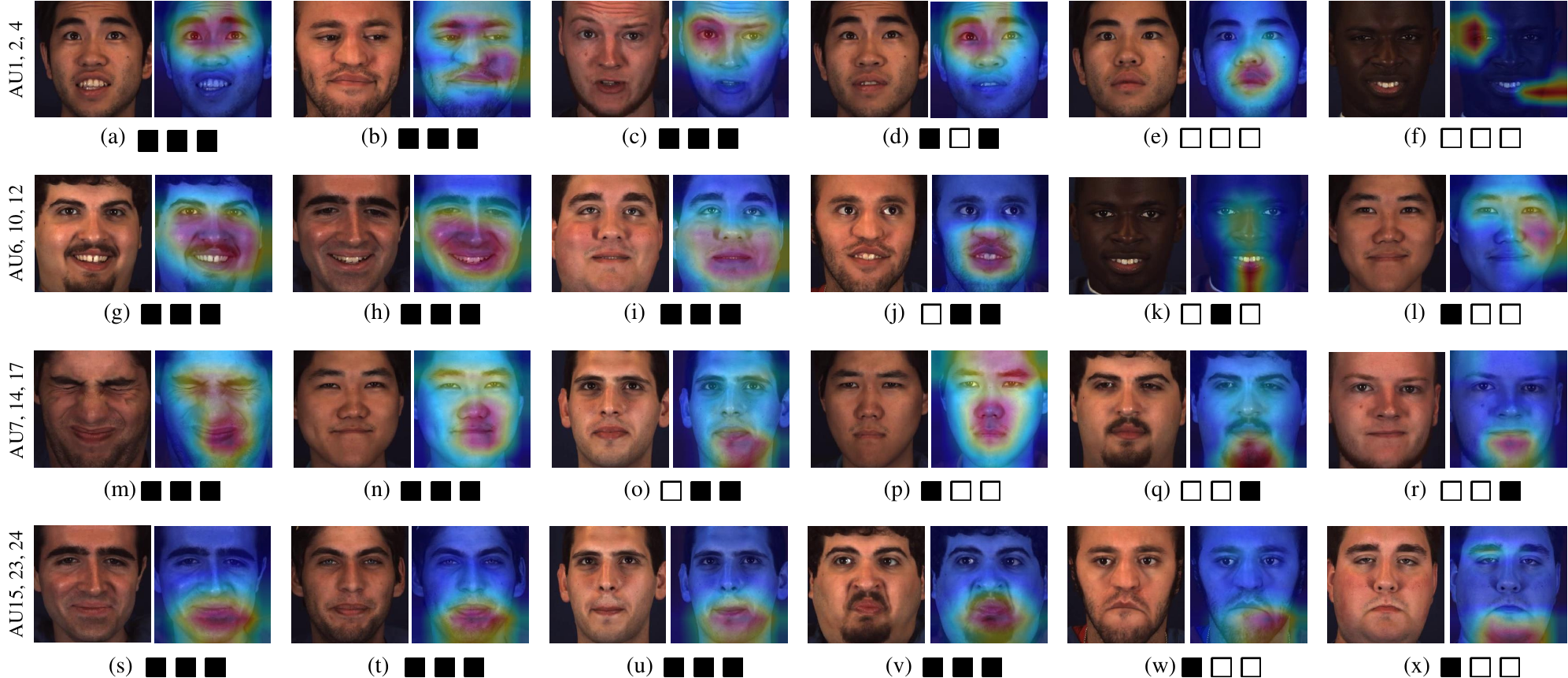}
	\caption{
		Visualization of attention maps of CLP and the AU detection results in BP4D dataset. $\blacksquare$ denotes the successful and $\square$ means the failed AU predictions.
	}
	\label{fig:bp4d}
\end{figure*}

\begin{figure*}[htb]
	\centering
	\includegraphics[width=0.8\linewidth]{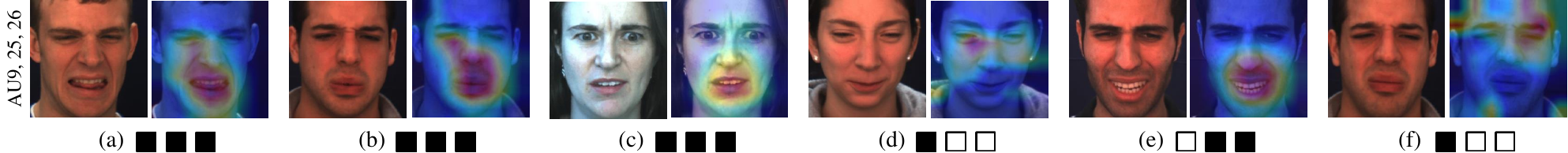}
	\caption{
		Visualization of attention maps of CLP and the AU detection results in DISFA dataset. $\blacksquare$  denotes the successful and $\square$ means the failed AU predictions.
	}
	\label{fig:disfa}
\end{figure*}

\subsubsection{Comparison on the facial expression recognition (FER) datasets:}  
We show the FER results in Tab.~\ref{tab:rafdb_tae_test} and Tab.~\ref{tab:affectNet_tae_test}.
For the comparisons in Tab.~\ref{tab:rafdb_tae_test} and Tab.~\ref{tab:affectNet_tae_test}, we compare our proposed CLP with several representative \textbf{supervised FER methods}, including AlexNet~\cite{krizhevsky2017imagenet}, VGG16~\cite{simonyan2014very}, gACNN \cite{li2018occlusion} and VTFF \cite{9585378}. In addition, we also compare CLP with the representative \textbf{self-supervised representation learning methods}, including DeformAE~\cite{shu2018deforming}, Fab-Net \cite{wiles2018self} and TAE \cite{li2020learning}. For the compared method, \textit{gACNN} was designed to learn the region-specific features for robust FER. \textit{VTFF} combines shadow and deep features to further enrich the representation of the visual words. It then models the relationship between these visual words with the transformer mechanism.

As verified in Tab.~\ref{tab:rafdb_tae_test} and Tab.~\ref{tab:affectNet_tae_test}, CLP consistently outperforms other self-supervised methods. However, CLP illustrates obvious performance degradation when compared with the state-of-the-art supervised FER methods, e.g., VTFF \cite{9585378}. The comparisons on RAF-DB and AffectNet datasets indicate the superiority of CLP is not so obvious on FER as that on AU detection. We conclude the main reasons as two-fold: (1) The FER datasets, e.g., RAF-DB and AffectNet, already consist of rather sufficient and diverse images. Notably, there are approximately 280,000 images for training in the AffectNet dataset. Thus, the supervised FER model is capable of learning more discriminative representations. As a comparison, AU detection datasets usually contain few subjects, e.g., 41 subjects in BP4D and 27 subjects in DISFA. Thus, AU detection task can benefit more from our CLP-pre-trained models. (2) Our proposed CLP is reinforced to learn the frame-wisely discriminative representations for AU detection. However, this is not a prior for FER because the faces in a video clip may share the same facial expression category.

\subsection{Ablation study}
\label{sec:ablation_study}
We conduct ablation studies to investigate the contributions of $\mathcal{L}_{tcl}$ and $\mathcal{L}_{cir}$ in CLP, the influence of the weight $\lambda$ for the triplets in Equ.~\ref{eq:triplet_loss}, and how the size of the dictionary $|C|$ influences the distinctiveness of the encoded AU representations in CLP. The results are shown in Tab.~\ref{tab:ablationstudy}. The observations are three-fold:  (1) $\mathcal{L}_{cir}$ is the cutting edge term in CLP. It is because $\mathcal{L}_{cir}$ can learn the intra-image invariance and the cross-video consistent AU representation simultaneously. It can reduce the influence of AU-independent nuisances such as identity and thus eliminate the person-specific features. (2) Increasing the nearest dictionary size to $32768$ gives us most discriminative AU representations in the CIR learning part. Increasing or decreasing the dictionary size will lead to sub-optimal solutions. It suggests the facial AUs are fine-grained and a large dictionary may better sample the underlying huge, high-dimensional AU representation space.
(3) $\lambda_j$ should be monotonically descending with the increase of the $j$ which denotes the temporal intervals between the anchor and the positive frames within a short video clip. Obviously, $\lambda_j = 1$ obtains inferior AU detection performance. CLP obtains the best AU detection performance with $\lambda_j = 1 / \sqrt{j}$ compared with other choices. 

We visualize the learned features of CLP and CLP without $\mathcal{L}_{cir}$ (CLP (\textit{w/o CIR})) to qualitatively investigate the effectiveness of CIR in Fig.~\ref{fig:bp4d_feature_visualize}, Fig.~\ref{fig:disfa_feature_visualize}, Fig.~\ref{fig:gft_feature_visualize}. 
Furthermore, we visualize the features of the supervised ResNet-34 model for a comprehensive comparison. For each dataset, we randomly sampled 30 images from 10 subjects for illustration. 
The top rows in the preceding two columns in Fig.~\ref{fig:bp4d_feature_visualize}, ~\ref{fig:disfa_feature_visualize}, ~\ref{fig:gft_feature_visualize} show that the CLP-learned AU representation are more separable than the representations of CLP (\textit{w/o CIR}). Besides, the bottom rows in the preceding two columns illustrate that the CLP-learned features are more scattered and dispersed among various subjects, indicating the features contain less subject-dependent components and are more invariant to subjects.
Compared with the supervised baseline, our proposed CLP shows a bit more discriminative w.r.t AU and more dispersed w.r.t subjects. It suggests the CLP-learned AU representations are more subject-invariant than the supervised baseline.

Besides, we show the successful \& failure AU detection cases on BP4D and DISFA dataset in Fig. \ref{fig:bp4d} and Fig. \ref{fig:disfa}, respectively. The AUs are divided into several groups for illustration convenience. $\blacksquare$  denotes the successful and $\square$ means the failed AU predictions. We have three observations. Firstly, CLP is capable of recognizing the AUs across different subjects. Secondly, CLP might fail for the rarely seen samples (Fig. \ref{fig:bp4d}(f), (k)), the low-intentity AUs (Fig. \ref{fig:bp4d}(e), (l), (q), (w)).  Thirdly, CLP may fail in the under-represented AUs. e.g., AU9 in DISFA dataset in Fig. \ref{fig:disfa} (e). It can be explained that the training datasets are multi-label-based, and several specific AU categories are highly imbalanced, e.g., merely 5.4\% AU9 are activated in the DISFA dataset. We will explore how to enhance the detection performance for the rarely seen samples and AUs in future work.

\section{Conclusion}
This paper present an approach to contrastively learn the person-independent AU representation from unlabeled facial videos. The goal of our proposed CLP is to learn the AU representations that are discriminative within a short video clip and consistent between different identities that show similar AUs. Experimental results show CLP outperforms the state-of-the-art self-supervised learning methods and considerably closes the performance gap between the self-supervised and supervised AU detection approaches. Qualitative feature visualization results have verified our hypothesis on the learned AU representations. For further work, we plan to incorporate the transformer mechanism and perceive the structural information among the AUs in the self-supervised paradigm.

\section{Acknowledgment}
This work was supported by the National Key R\&D Program of China under contract NO.2018AAA0102402, the National Natural Science Foundation of China (62102180), the Natural Science Foundation of Jiangsu Province (BK20210329), Shuangchuang Program of Jiangsu Province(JSSCBS20210210).

\bibliographystyle{IEEEtran} 
\bibliography{reference}

\begin{IEEEbiography}[{\includegraphics[width=1in,height=1.25in,clip,keepaspectratio]{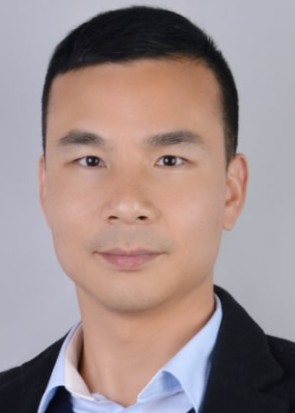}}]
{Yong Li} Received a Ph.D. degree in computer science from Institute of Computing Technology (ICT), Chinese Academy of Science (CAS), Beijing, in July, 2020. He has been an assistant professor at School of Computer Science and Engineering, Nanjing University of Science and Technology since 2020. His research interests include deep learning, self-supervised learning and affective computing.
\end{IEEEbiography}

\begin{IEEEbiography}[{\includegraphics[width=1in,height=1.25in,clip,keepaspectratio]{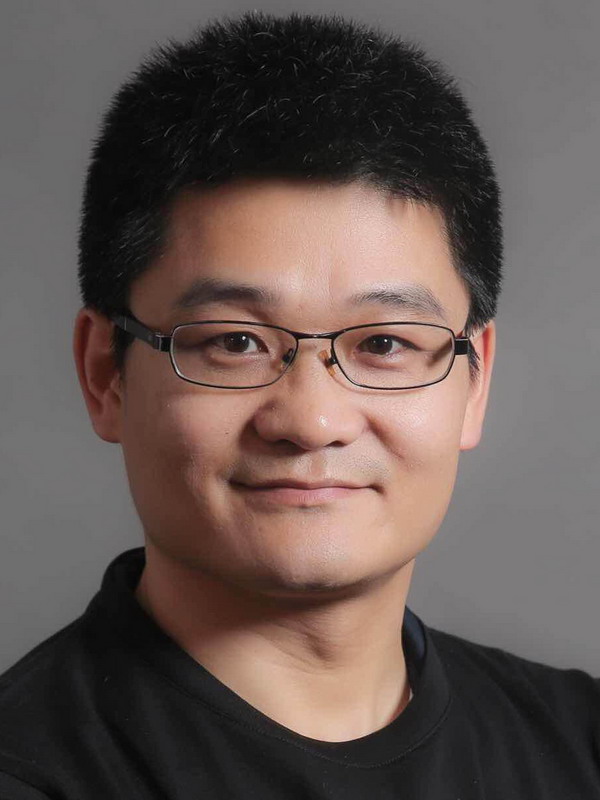}}]
{Shiguang Shan} received M.S. degree in computer science from the Harbin Institute of Technology, Harbin, China, in 1999, and Ph.D. degree in computer science from the Institute of Computing Technology (ICT), Chinese Academy of Sciences (CAS), Beijing, China, in 2004. He joined ICT, CAS in 2002 and has been a Professor since 2010. He is now the deputy director of the Key Lab of Intelligent Information Processing of CAS. His research interests cover computer vision, pattern recognition, and machine learning. He has published more than 300 papers in refereed journals and proceedings in the areas of computer vision and pattern recognition. He has served as Area Chair for many international conferences including CVPR, ICCV, IJCAI, AAAI, ICPR, ACCV, FG, etc.. He is Associate Editors of several international journals including IEEE Trans. on Image Processing, Computer Vision and Image Understanding, Neurocomputing, and Pattern Recognition Letters. He is a recipient of the China's State Natural Science Award in 2015, and the China’s State S\&T Progress Award in 2005 for his research work. 
\end{IEEEbiography}

\end{document}